\documentclass[times,review]{elsarticle}
\usepackage{float}

\usepackage{dib}
\usepackage{framed,multirow}
\usepackage{comment}
\usepackage{booktabs}

\usepackage{amssymb}
\usepackage{latexsym}

\usepackage{url}
\usepackage{graphicx}
\usepackage{xcolor}
\definecolor{newcolor}{rgb}{.8,.349,.1}

\usepackage{longtable}
\usepackage[colorlinks]{hyperref}
\graphicspath{ {./Visualisations/} }
\journal{Data in Brief}

\begin{document}

\verso{R.S. Kiziltepe \textit{et. al}}

\begin{frontmatter}

%\dochead{Data Article}

\title{An Annotated Video Dataset for Computing Video Memorability\tnoteref{tnote1}}%

\author[1]{Rukiye Savran \snm{Kiziltepe}}  
\author[2]{Lorin \snm{Sweeney}}
\author[3]{Mihai Gabriel \snm{Constantin}}
\author[1]{Faiyaz \snm{Doctor}}
\author[1]{Alba \snm{Garc\'ia Seco de Herrera}}
\author[4]{Claire-Hélène \snm{Demarty}}
\author[2]{Graham \snm{Healy}}
\author[3]{Bogdan \snm{Ionescu}}
\author[2]{Alan F. \snm{Smeaton}\corref{cor1}}
\cortext[cor1]{Corresponding author.
%:   Tel.: +0-000-000-0000;  
%  fax: +0-000-000-0000;
}

\ead{Alan.Smeaton@DCU.ie}

%%Affiliations
\address[1]{University of Essex, UK}
\address[2]{Insight Centre for Data Analytics, Dublin City University, Glasnevin, Dublin 9, Ireland}
\address[3]{University Politehnica of Bucharest, Romania}
\address[4]{InterDigital, R\&I, France}

\begin{abstract}
%%%%
Using a collection of publicly available links to short form video clips of an average of 6 seconds duration each, 1,275 users manually annotated each video multiple times to indicate both long-term and short-term memorability of the videos.  The annotations were gathered as part of an online memory game and measured a participant's ability to recall having seen the video previously when shown a collection of videos. The recognition tasks were performed on videos seen within the previous few minutes for short-term memorability and within the previous 24 to 72 hours for long-term memorability.  Data includes the reaction times for each recognition of each video.  Associated with each video are text descriptions (captions) as well as a collection of image-level features applied to 3 frames extracted from each video (start, middle and end). Video-level features are also provided.  The dataset was used in the Video Memorability task as part of the MediaEval benchmark in 2020.
%\noindent[The Abstract should describe the data collection process, the analysis
%performed, the data, and their reuse potential. It should not provide
%conclusions or interpretive insights. 

%\noindent\textbf{Tip:} do not use words such as
%`study', `results', and `conclusions' because a data article should be
%describing your data only.  Min 100 words - Max 500 words.]
%%%%
\end{abstract}

\begin{keyword}
%% Keywords
%[Include 4-8 keywords (or phrases) to facilitate others finding your
%article online. 
%\noindent\textbf{Tip:} Try Google Scholar to find which terms are most common in your
%field. In biomedical fields, MeSH terms are a good 'common vocabulary'
%to draw from]
\KWD Video memorability\sep Machine learning\sep Human memory\sep MediaEval Benchmark
\end{keyword}

\end{frontmatter}

{\fontsize{7.5pt}{9pt}\selectfont
%%%
\noindent\textbf{Specifications Table} 

\begin{longtable}{p{33mm}p{94mm}}

\endhead

\endfoot
Subject                & Computer Vision and Pattern Recognition % [Please select one CATEGORY for your manuscript from the list available at:\break \href{https://www.elsevier.com/__data/assets/excel_doc/0012/736977/% DIB-categories.xlsx}{DIB categories}.]
\\
                         
Specific subject area  & Ground truth data (videos, video features plus annotations) needed to build and train systems for the automatic computation of the  memorability of short video clips %[Briefly describe the narrower subject area. Max 150 characters]
\\

Type of data           & Text files (csv) \\
                        % [List the type(s) of data this article describes. 
                        %  Simply delete from this list as appropriate:] 
                        %  Table\newline
                        %  Image\newline
                        %  Chart\newline
                        %  Graph\newline
                        %  Figure\newline
                        %  [Any other type not listed- please specify]\\             
%\clearpage
How data were acquired & Raw videos are already publicly available online. Low level features were extracted automatically from videos and annotation data was collected through crowdsourcing  using a video memorability game with the participation of both volunteers  and paid workers on Amazon Mechanical Turk. \\ %\chelene{Should we also mention that we used crowdsourcing?}
                        %   [State how the data were acquired: E.g. Microscope,  
                        %  SEM, NMR, mass spectrometry, survey* etc.\newline
                        %  Instruments: E.g. hardware, software, program\newline
                        %  Make and model and of the instruments used:\newline

                        %  {\fontsize{7pt}{8pt}\selectfont
                        %  *\,if you conducted a survey you must submit a copy of the 
                        %  survey(s) used (either provide these as supplementary material 
                        %  file or provide a URL link to the survey 
                        %  in this section of the table). 
                        %  If the survey is not written in English, 
                        %  please provide an English-language translation.}]\\
                         
Data format            & Raw \newline
                         Analyzed\\
                        % [List your data format(s). Note, unless you are describing secondary data, 
                        %  all raw data must be provided (either with this data article or linked to a repository). 
                        %  Simply delete from this list as appropriate:]\newline
                        %  Raw\newline
                        %  Analyzed\newline
                        %  Filtered\newline
                        % [Any other format not listed- please specify]\\
                         
Parameters for         
data\newline 
collection             & The maximum false alarm rate (short-term): 30\% \newline
                        The maximum false alarm rate (long-term): 40\% \newline
                        The minimum recognition rate of vigilance fillers (short-term): 70\%  \newline
                        The minimum recognition rate (long-term): 15\% \newline
                        The false alarm rate must be lower than the recognition rate (long-term).\\  
%[Provide a brief description of which conditions were considered for data collection. Max 400characters]\\  

Description of          
data\newline 

                        %[Provide a brief description of how these data were collected. Max 600 characters]\\
collection             & 1,500 short videos  selected from the Vimeo Creative Commons (V3C1) dataset and used in the TRECVid 2019 Video-to-Text task  were  divided into three non-overlapping subsets: training, development, and testing. Multiple manual memorability annotations for each video were collected via a video memorability game, which displays a series of short videos and requires  users to press the spacebar when they recall a video previously seen by them. The game consists of two parts: in the first part where  videos are repeated within a few minutes, the user interaction with a repeated video was collected to calculate short-term memorability scores. The second part took place between 24  and 72 hours after initial viewing of videos, and this time the participants' responses to  previously seen videos from the first part were collected to acquire  long-term memorability scores. After analysing the collected annotations, the short-term and the long-term memorability scores of each video were calculated as a percentage of correctly recalled videos, respectively. Each video memorabiity annotation is accompanied by the video timepoint offsets at which it was recalled by users,  response times of the users, the key pressed when watching each video, and textual captions describing each video from the TRECVid benchmark. The Media Memorability 2020 dataset is included here with memorability annotations on  590 videos as part of the training set and 410 additional videos as part of the development set. In this dataset we provide memorability annotations for the development and training set videos but not the test set as this is used in  future MediaEval memorability benchmark  tasks.\\
                         
Data source location   & Primary data sources: TRECVid 2019 Video-to-Text dataset \cite{TRECVID2019}, available from: \url{https://www-nlpir.nist.gov/projects/trecvid/trecvid.data.html} \newline
Institution: National Institute of Standards and Technology (NIST) \newline
City: Gaithersburg \newline
Country: USA \\

                        % [Fill in the information available, and delete from this list as appropriate:\newline
                        %  
                         % Latitude and longitude (and GPS coordinates, if possible) for collected samples/data:\newline
                        %  If you are describing secondary data, you are required to provide a list of 
                        %  the primary data sources used in the section.\newline

                        %  Primary data sources:  ]\\
                         
%\hypertarget{target1}
{Data accessibility}   & Repository name: Figshare \newline
%Data identification number: {to be added} \newline
Direct URL to data: \url{https://doi.org/10.6084/m9.figshare.15105867.v2}
\\
&Source code used to process this is adapted from ~\cite{CDD2019} and is available at \url{https://github.com/InterDigitalInc/VideoMemAnnotationProtocol/}\\
       
Related                 
research\newline
article                &   A. Garc\'ia Seco  de  Herrera,  R.  Savran  Kiziltepe,  J.  Chamberlain,  M.  G.  Constantin,  C.-H.  Demarty,  F.  Doctor, B. Ionescu, A. F. Smeaton\newline 
Overview of MediaEval 2020 Predicting Media Memorability Task: What Makes a Video Memorable?\newline 
MediaEval Workshop, online 14-15 December, 2020~\cite{GSC2020}\newline
DOI: \url{}
\\
                        % [If your data article is related to a research article - \textbf{especially 
                        %  if it is a co-submission} - please cite your associated research 
                        %  article here. Authors should only list \textbf{one article}.\newline

                        %  Authors' names\newline
                        %  Title\newline
                        %  Journal\newline
                        %  DOI: \textbf{OR} for co-submission manuscripts `In Press'\newline

                        %  \textbf{For example, for a direct submission:}\newline

                        %  J. van der Geer, J.A.J. Hanraads, R.A. Lupton, The art of writing a scientific article, 
                        %  J. Sci. Commun. 163 (2010) 51-59. https://doi.org/10.1016/j.Sc.2010.00372\newline

                        %  \textbf{Or, for a co-submission (when your related research article has not yet published):}\newline

                        %  J. van der Geer, J.A.J. Hanraads, R.A. Lupton, The art of writing a 
                        %  scientific article, J. Sci. Commun. In Press.\newline

                        %  \textbf{Or, if your data article is not directly related to a research article, 
                        %  please delete this last row of the table.}]
\end{longtable}
}
%%%            

\section*{Value of the Data}

\begin{itemize}
\itemsep=0pt
\parsep=0pt
\item %Your first bullet point must explain why these data are useful or important? 
Media platforms such as social networks, media advertisement, information retrieval and recommendation systems deal with exponentially growing data day after day. Enhancing the relevance of multimedia data -- including video --  in our everyday lives requires new ways to analyse, index and organise such data. In particular it requires us to be able to discover, find and retrieve  digital content like video clips and that means automatically analysing video so that it can be found. Much work in the computer vision community has concentrated on analysing video in terms of its content, identifying objects in the video or activities taking place in the video but video has other characteristics  such as aesthetics or interestingness or memorability.  Video memorability refers to how easy it is for a person to remember seing a video and video  memorability can be regarded as useful for a system to make a choice between competing videos on which video to present to a user when that user is searching for video clips.  Video memorability will also be useful in areas like online advertising or video production where the memorability of a video clip will be important. The data provided here can be used to train a machine learning system to automatically calculate the likely memorability of a short form video clip. 
 
\item %Your second bullet point must explain who can benefit from these data?
Researchers will find this data interesting if they work in the areas of human perception and scene understanding, such as image and video interestingness, memorability, attractiveness, aesthetics prediction, event detection, multimedia affect and perceptual analysis, multimedia content analysis, or machine learning.   

\item %Your third point bullet must explain how these data might be used/reused for further insights and/or development of experiments.
The dataset provides links to publicly available short form video clips, each of 6 seconds duration, features which describe those videos and annotations as to the memorability of those videos. This is all the data needed to train and evaluate the accuracy of machine learning classifier to predict video memorability.

%\item In the next three points you may like to explain how these data could potentially make an impact on society and highlight any other additional value of these data.
\item A huge amount of video material is now available to us at our fingertips, including from video sharing platforms like YouTube and Vimeo, video streaming platforms like Netflix and Amazon Prime, videos shared on social media platforms and even the video clips we ourselves generate on our smartphones. Unlike searching text documents on the WWW, searching through all this video content in order to find a clip you may have seen previously or a clip you think might exist but you are not sure and you would like to find it, such information search is not currently supported.  Eventually technology companies will catch up with the growth in the amount of available video content and as they do, the intrinsic memorability of a video clip will be a characteristic of a video clip that will be important in deciding whether to retrieve it for a user.  This means that video search will give us search results which will be with the better, more memorable videos more highly ranked.

\item The specific use cases of creating video commercials or creating educational content requires videos which people will remember. Because the impact of different forms of visual multimedia content -- images or videos -- on human memory is unequal, the capability of predicting or computing the likely memorability  of a clip of video content is obviously of high importance for professionals in the fields of advertising and education.

\item Beyond advertising and educational applications, other areas such as filmmaking will find use for methods which calculate the memorability of video clips.  We may see film and documentary makers creating videos in such a way that the key moments in a movie or documentary will be created in ways so as to maximise their likely memorability by the viewer and that opens up new ways for creating video material.

\end{itemize}

\section{Data Description}

\begin{figure}[htb]
\centering
\includegraphics[width=0.65\textwidth]{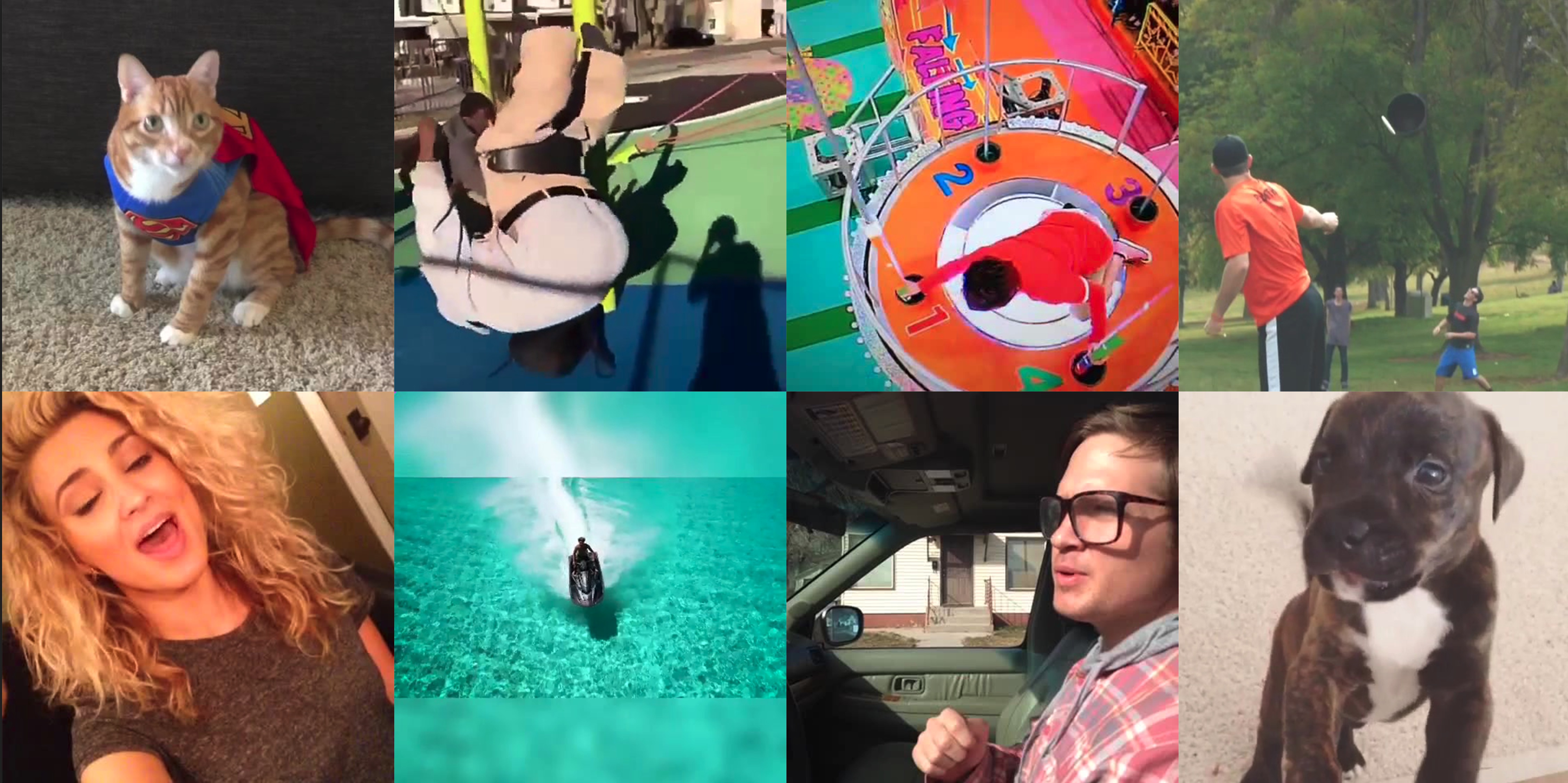}
\caption{A sample of frames from some of the videos in the \textit{TRECVid 2019 Video-to-Text dataset.}}
\label{fg:videos}
\end{figure}

The Media Memorability 2020 dataset contains a subset of short videos selected from the TRECVid 2019 Video-to-Text dataset~\cite{TRECVID2019} and a sample of frames from some of these is shown in Figure~\ref{fg:videos}. The dataset  contains links to, as well as features describing and annotations on, 590 videos as part of the training set and 410  videos as part of development set. It also contains links to, and features describing, 500 videos used as test videos for the MediaEval Video Memorability benchmark in 2020.

%-------------------------------------
%-------------------------------------

\begin{figure}
\centering
\includegraphics[width=0.95\textwidth]{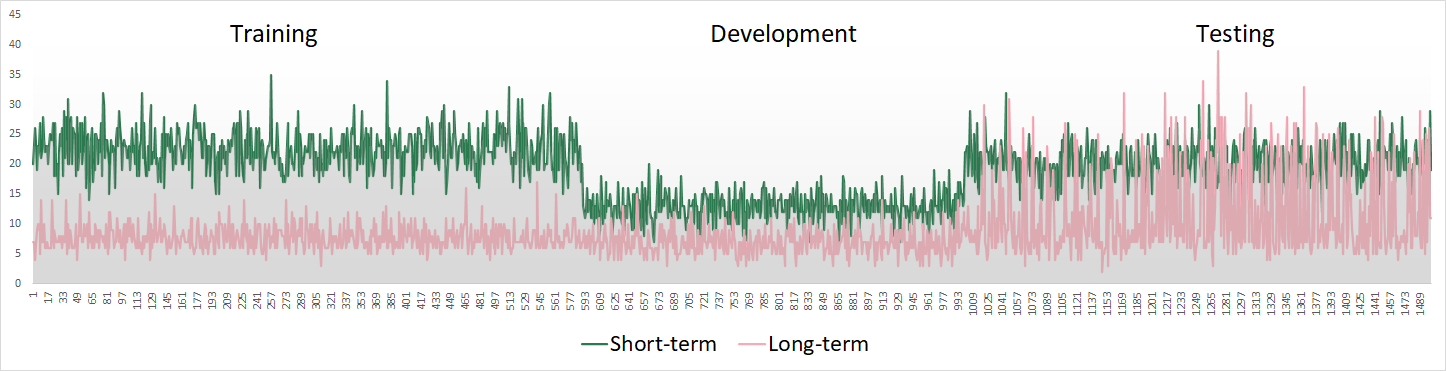}
\caption{The number of annotations in the training, development, and test sets.}
\label{fg:num_annotations}
\end{figure}

Each video in the training and development sets is distributed with both short-term and long-term memorability ground truth scores and several automatically calculated features. The collected annotations for each video are also published along with the overall memorability scores. We collected a minimum of 14 and a mean of 22 annotations in the short-term memorability step, and a  minimum of 3 and a mean of 7 annotations in the long-term memorability step for the training set. The development set has similar annotation numbers in the long-term step with a minimum of 3 and a mean of 7 annotations, however, the number of annotations for the development set in the short-term step is lower than in the training set with a minimum of 6 and a mean of 12.    Figure~\ref{fg:num_annotations} shows the distribution of the number of annotations for short-term and long-term memorability.  We will continue improving and update the development set with more annotations in the near future.

Five files are released for each of the training and development sets as presented in Table~\ref{tb:annotations}.

%-------------------------------------
%-------------------------------------

%-------------------------------------
{\vskip12pt}
{\fontsize{7.5pt}{9pt}\selectfont
\begin{table}[ht]
    \centering
    \begin{tabular}{ll}
    \toprule 
        Training Set & Development Set\\
        \midrule
video\_urls.csv & dev\_video\_urls.csv \\

short\_term\_annotations.csv & dev\_short\_term\_annotations.csv \\

long\_term\_annotations.csv & dev\_long\_term\_annotations.csv \\

scores.csv & dev\_scores.csv \\

text\_descriptions.csv & dev\_text\_descriptions.csv \\
\bottomrule
    \end{tabular}
    \caption{Text files in the training and development sets of the MediaEval2020 Predicting Media Memorability dataset}
    \label{tb:annotations}
\end{table}
}
%-------------------------------------

\noindent 
The dataset also contains the same features and structure in the training and development sets. Table~\ref{table3} presents the text files with the features and their descriptions.
\pagebreak
{\vskip12pt}
{\fontsize{7.5pt}{9pt}\selectfont
\begin{longtable}{p{15mm}p{25mm}p{60mm}}
\caption{The text files in the training and development sets of the MediaEval2020 Predicting Media Memorability dataset}\\
\hline
Text File & Feature Name & Description \\
\hline
\multicolumn{ 1}{l}{Video URLs} &  video\_id &  the unique video id \\

\multicolumn{ 1}{l}{} & video\_url &  the video url \\
\hline
\multicolumn{ 1}{l}{Short-term annotations} &  video\_id &  the unique video id \\

\multicolumn{ 1}{l}{} & video\_url &  the video url \\

\multicolumn{ 1}{l}{} &   user\_id &  the id number of the user performing the annotation \\

\multicolumn{ 1}{l}{} &         rt &  response time in milliseconds for the second occurrence of the video ( -1 for no response given by the user) \\

\multicolumn{ 1}{l}{} & key\_press &  the key code pressed by the user for the second occurrence of the video (32 is for spacebar, -1 for no response) \\

\multicolumn{ 1}{l}{} & video\_position\_first &  the position of video seen first time in the current stream (1-180) \\

\multicolumn{ 1}{l}{} & video\_position\_second &  the position of video seen second time in the current stream (1-180) \\

\multicolumn{ 1}{l}{} &    correct &  1 is for the correct response 0 is for incorrect response \\
\hline
\multicolumn{ 1}{l}{Long-term annotations} &  video\_id &  the unique video id \\

\multicolumn{ 1}{l}{} & video\_url &  the video url \\

\multicolumn{ 1}{l}{} &   user\_id &  the id number of the user performing the annotation \\

\multicolumn{ 1}{l}{} &         rt &  response time in milliseconds for the occurrence of the video ( -1 for no response given by the user) \\

\multicolumn{ 1}{l}{} & key\_press &  the key code pressed by the user for the second occurrence of the video (32 is for spacebar, -1 for no response) \\

\multicolumn{ 1}{l}{} & video\_position &  the position of target video seen in the current stream (1-180) \\

\multicolumn{ 1}{l}{} &    correct &  1 is for the correct response 0 is for incorrect response \\
\hline
\multicolumn{ 1}{l}{Text Descriptions} &  video\_id &  the unique video id \\

\multicolumn{ 1}{l}{} & video\_url &  the video url \\

\multicolumn{ 1}{l}{} & description &  text description for the video \\
\hline
\multicolumn{ 1}{l}{Scores} &  video\_id &  the unique video id \\

\multicolumn{ 1}{l}{} & video\_url &  the video url \\

\multicolumn{ 1}{l}{} &     ann\_1 &  the number of annotations for short-term memorability \\

\multicolumn{ 1}{l}{} &     ann\_2 &  the number of annotations for long-term memorability \\

\multicolumn{ 1}{l}{} & part\_1\_scores &  short-term memorability score \\

\multicolumn{ 1}{l}{} & part\_2\_scores &  long-term memorability score \\
\hline
\label{table3}
\end{longtable}
}

\noindent 
Additional pre-computed features are provided in individual folders per feature type and in individual csv files per sample, which are available in the  data repository. There are seven folders containing the seven features for each of the training, development and test sets as follows:

{\fontsize{7.5pt}{9pt}\selectfont
\begin{itemize}
\itemsep=0pt
\parsep=0pt
\item AlexNetFC7 (image-level feature)~\cite{krizhevsky2012imagenet}
\item HOG (image-level feature)~\cite{dalal2005histograms}
\item HSVHist (image-level feature)
\item RGBHist (image-level feature)
\item LBP (image-level feature)~\cite{ojala2002multiresolution}
\item VGGFC7 (image-level feature)~\cite{simonyan2014very}
\item C3D (video-level feature)~\cite{tran2015learning}
\end{itemize}
}

\noindent 
For  image-level features we extract features from 3 frames for each video, each one in an individual file, where the filenames are composed as follows: $<$video\_id$>$-$<$frame\_no$>$.csv.
The 3 frames per  video represent the first, the middle and the last frames in the video clip. For example, for video\_id 8 we extract the following AlexNet feature-files):

{\fontsize{7.5pt}{9pt}\selectfont
\begin{itemize}
\itemsep=0pt
\parsep=0pt
\item AlexNetFC7/00008-000.csv : AlexNetFC7 feature for video\_id = 8, frame\_no = 0 (first frame)
\item AlexNetFC7/00008-098.csv : AlexNetFC7 feature for video\_id = 8, frame\_no = 98 (middle frame)
\item AlexNetFC7/00008-195.csv : AlexNetFC7 feature for video\_id = 8, frame\_no = 195 (last frame)
\end{itemize}
}

\noindent 
For  video-level features we extract 1 feature for each video, where the filenames are composed as follows: $<$video\_id$>$.mp4.csv. Using the same video\_id 8 as an example, we extract the following C3D feature-file:

{\fontsize{7.5pt}{9pt}\selectfont
\begin{itemize}
\itemsep=0pt
\parsep=0pt
\item C3D/00008.mp4.csv : C3D features for video\_id = 8
%\item  ...
\end{itemize}
}

\begin{figure}[ht]
    \centering
\includegraphics[width=\textwidth]{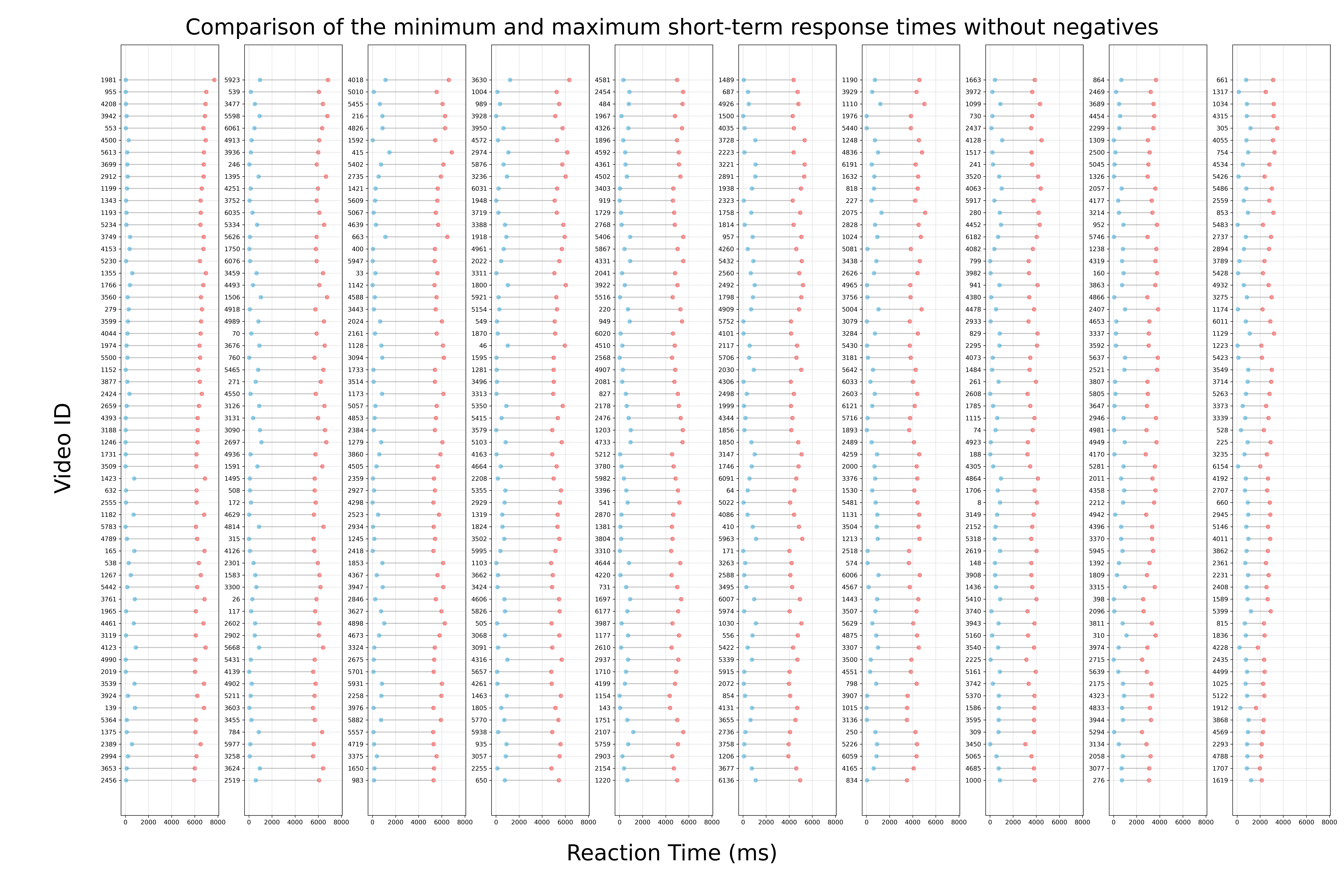}
\caption{Comparison of the short-term minimum and maximum reaction times for 510 videos. \label{fig:reachtiontime-short}}
\end{figure}

\begin{figure}
\includegraphics[width=\textwidth]{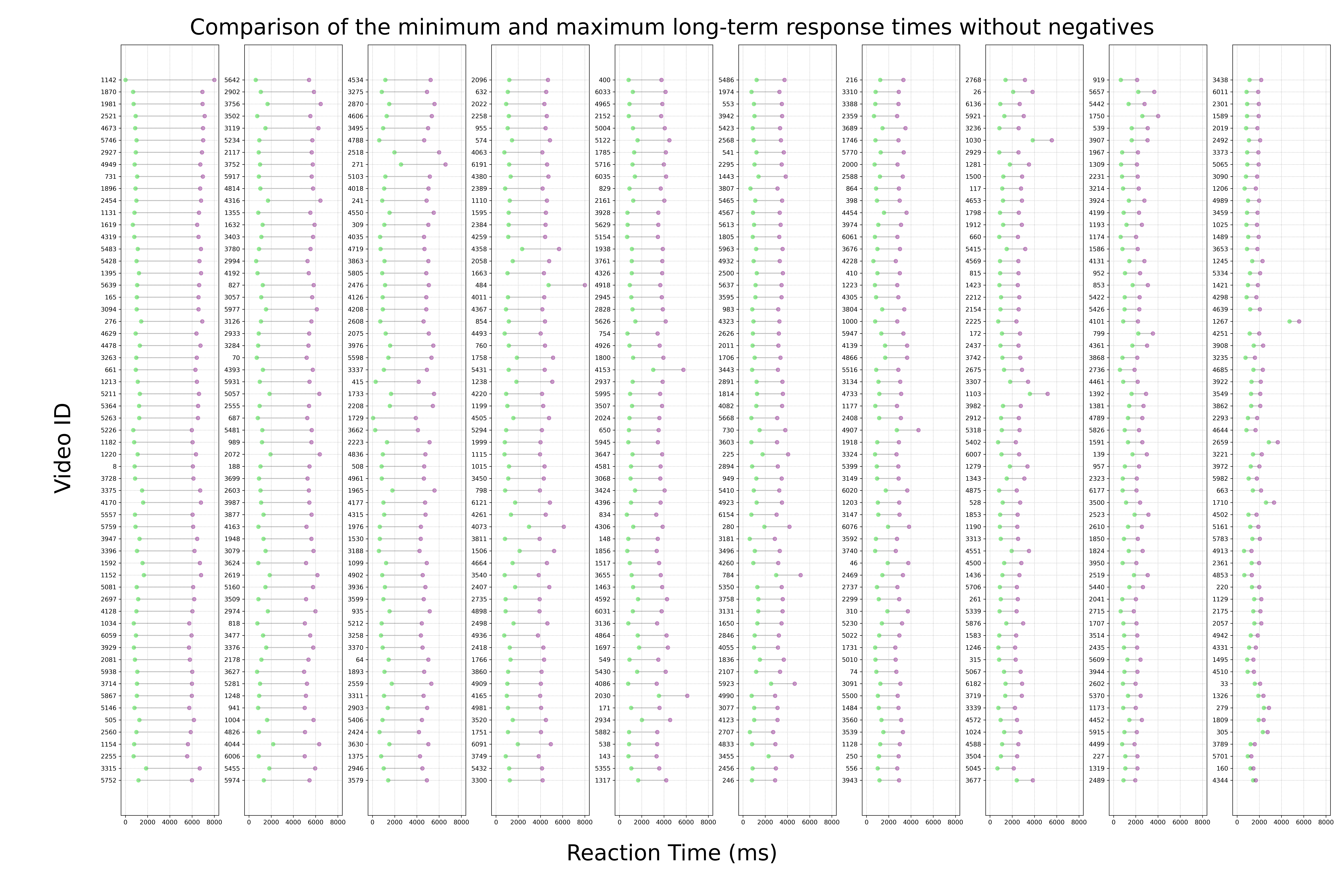}
\caption{Comparison of the long-term minimum and maximum reaction times for 510 videos.\label{fig:reachtiontime-long}}
\end{figure}

\noindent 
Figure~\ref{fig:reachtiontime-short} shows the minimum and maximum reaction times for the annotations for short-term memorability for each of the 590 videos in the training set while Figure~\ref{fig:reachtiontime-long} shows the same for long-term memorability.  The figures reads from left to right, with each  column being the vertical continuation of the preceding column. Reaction times are sorted greatest to shortest difference between minimum and maximum reaction time, x-axis is the reaction time in milliseconds, numbers on the y-axes refer to video\_id. The figure illustrates a large range of min-to-max reaction times, those appearing later in the graph (rightmost column, towards the bottom) appear to be universally memorable to all annotators while those at the other end of the graph are memorable to some as soon as video playback commences, and less memorable to others.  The positioning of the blue dots in the graph indicates that all videos have at least some annotators who remember the video early during the playback, in many instances almost as soon as video playback commences.  The differences between short- and long-term memorability annotations indicate long-term recall happens sooner, i.e., earlier during video playback.

\vspace*{\floatsep}

\section{Experimental Design, Materials and Methods}

Each video  has  two associated scores of memorability that refer to its probability to be remembered after two different durations of memory retention. Memorability has been measured using recognition tests, i.e., through an objective measure, a few minutes after the memorisation of the videos (short-term), and then 24 to 72 hours later (long-term). 

The ground truth dataset was collected  using a video memorability game protocol proposed by Cohendet et al.~\cite{CDD2019}. In a first step (short-term memorization), participants  watched 180 videos, among which 40 target videos are repeated after a few minutes to collect short-term memorability labels. The task is basically to press the space bar once a participant recognises a previously seen video, which enables to determine videos recognised and not recognised by them. As for filler videos in the first step, 60 non-vigilance filler videos are displayed once. 20 vigilance filler videos are repeated after a few seconds to check  participants’ attention to the task. 

After between 24 and 72 hours, the same participants attend the second step for collecting long-term memorability labels. During this second step, they each watch 120 videos comprised of 40 target videos chosen randomly from among non-vigilance fillers from the first step and 80 fillers selected randomly from new videos which are displayed to measure long-term memorability scores for those target videos. 

Both short-term and long-term memorability scores are calculated as the percentage of correct recognitions for each video, by the participants.

The experimental protocol was written in PhP and JavaScript (a  modified version of the JavaScript library in \cite{deLeeuw_2015_jspsych} was used) and interacts with a MySQl database. The interaction with Amazon mechanical turk was performed through JavaScript code. The optimisation problem for generating positions was written in Matlab. 

A participant could participate only once in the study.
The order of videos was randomly assigned, using an algorithm that randomly selects from among the last 1,000 least annotated videos, and which generates random positions from 45 to 100 videos (i.e., 4 to 9 minutes).

Several vigilance tests were  settled up upon the results on an in-lab test and only participants that met the controls were retained for the analysis:
\begin{enumerate}
    \item 20 vigilance fillers were added in the short-term step and we expected a recognition rate of those fillers of 70$\%$.
    \item a minimal recognition rate of $15\%$ in the long-term step.
    \item a maximal false alarm rate of $30\%$ for short-term and $40\%$ for long-term.
    \item a false alarm rate lower than the recognition rate for long-term.
\end{enumerate}

Two versions of the memorability game using three language options: English, Spanish and Turkish were published for different audiences and in different contexts. One was published on Amazon Mechanical Turk (AMT) and another  was issued for general use among an audience essentially made up of students. 
A total of 1,275 different users participated in the short-term memorability step while 602 participated in the long-term memorability step. Only about $48\%$ of the participants who completed the short-term step came back to participate in the long-term step.

\section*{Ethics Statement}

Institutional ethical approval for eliciting human participation for the memorability game was granted by the University of Essex, with protocol number ETH1920-1049. Anonymity of participants was maintained, and informed consent was obtained, and only data from consenting participants was used in constructing the memorability dataset. With respect to the use of Amazon Mechanical Turk (AMT), the design of the game and nature of information captured from  participants ensured security and confidentiality concerns were minimal.

\section*{CRediT Author Statement}

{\bf Rukiye Savran Kiziltepe}  Conceptualization, Methodology, Software Programming, software development, Validation, Formal analysis, Writing - Original Draft;
{\bf Lorin Sweeney:} Formal analysis, Writing - Original Draft, Visualization;
{\bf Mihai Gabriel Constantin:} Conceptualization, Methodology, Validation;
{\bf Faiyaz Doctor:} Methodology, Writing - Original Draft;
{\bf Alba Garc\'ia Seco de Herrera:} Conceptualization, Methodology, Validation, Writing - Original Draft;
{\bf Claire-Hélène Demarty:} Conceptualization, Methodology, Software Programming, software development, Validation, Writing - Original Draft, 
{\bf Graham Healy:} Methodology, Validation;
{\bf Bogdan Ionescu:} Conceptualization, Methodology, Validation;
{\bf Alan F. Smeaton:} Conceptualization, Methodology, Writing - Original Draft, Writing - Review \& Editing, Funding acquisition.

\begin{comment}
Conceptualization,  
Methodology, 
Software Programming, software development,  
Validation,  	
Formal analysis,  
Writing - Original Draft,  
Writing - Review & Editing,  
Visualization 
Funding acquisition 
\end{comment}

\section*{Declaration of Competing Interest}

The authors declare that they have no known competing
financial interests or personal relationships which have, or could be
perceived to have, influenced the work reported in this article.

\section*{Acknowledgements}
The work of Mihai Gabriel Constantin and Bogdan Ionescu was supported by the project AI4Media, A European Excellence Centre for Media, Society and Democracy, H2020 ICT-48-2020, grant \#951911.  
The work of Graham Healy, Alan Smeaton and Lorin Sweeney is partly supported by Science Foundation Ireland (SFI) under Grant Number SFI/12/RC/2289\_P2, co-funded by the European Regional Development Fund. 
The work of Rukiye Savran Kızıltepe is partially funded by the Turkish Ministry of National Education.
Funding for the annotation of videos was provided through an award from NIST  No. 60NANB19D155. 
We thank Cohendet et al.~\cite{CDD2019} for sharing their source code. 

%Acknowledgements should be inserted at the end of the paper, before the references, not as a footnote to the title. Use the unnumbered Acknowledgements Head style for the Acknowledgements heading.

% \section*{References}
\bibliographystyle{model1-num-names}
\bibliography{refs}

\begin{thebibliography}{9}
\expandafter\ifx\csname natexlab\endcsname\relax\def\natexlab#1{#1}\fi
\providecommand{\url}[1]{\texttt{#1}}
\providecommand{\href}[2]{#2}
\providecommand{\path}[1]{#1}
\providecommand{\DOIprefix}{doi:}
\providecommand{\ArXivprefix}{arXiv:}
\providecommand{\URLprefix}{URL: }
\providecommand{\Pubmedprefix}{pmid:}
\providecommand{\doi}[1]{\href{http://dx.doi.org/#1}{\path{#1}}}
\providecommand{\Pubmed}[1]{\href{pmid:#1}{\path{#1}}}
\providecommand{\bibinfo}[2]{#2}
\ifx\xfnm\relax \def\xfnm[#1]{\unskip,\space#1}\fi
%Type = Inproceedings
\bibitem[{Awad et~al.(2020)Awad, Butt, Curtis, Lee, Fiscus, Godil, Delgado,
  Zhang, Godard, Diduch, Smeaton, Graham, Kraaij, and Quenot}]{TRECVID2019}
\bibinfo{author}{G.~Awad}, \bibinfo{author}{A.~Butt},
  \bibinfo{author}{K.~Curtis}, \bibinfo{author}{Y.~Lee},
  \bibinfo{author}{J.~Fiscus}, \bibinfo{author}{A.~Godil},
  \bibinfo{author}{A.~Delgado}, \bibinfo{author}{J.~Zhang},
  \bibinfo{author}{E.~Godard}, \bibinfo{author}{L.~Diduch},
  \bibinfo{author}{A.~Smeaton}, \bibinfo{author}{Y.~Graham},
  \bibinfo{author}{W.~Kraaij}, \bibinfo{author}{G.~Quenot},
\newblock \bibinfo{title}{{TRECVID} 2019: An evaluation campaign to benchmark
  video activity detection, video captioning and matching, and video search \&
  retrieval},
\newblock in: \bibinfo{booktitle}{2019 TREC Video Retrieval Evaluation Notebook
  Papers and Slides}, \bibinfo{year}{2020}.
%Type = Inproceedings
\bibitem[{Cohendet et~al.(2019)Cohendet, Demarty, Duong, and
  Engilberge}]{CDD2019}
\bibinfo{author}{R.~Cohendet}, \bibinfo{author}{C.-H. Demarty},
  \bibinfo{author}{N.~Q. Duong}, \bibinfo{author}{M.~Engilberge},
\newblock \bibinfo{title}{{VideoMem}: Constructing, analyzing, predicting
  short-term and long-term video memorability},
\newblock in: \bibinfo{booktitle}{Proceedings of the {IEEE} International
  Conference on Computer Vision}, \bibinfo{year}{2019}, pp.
  \bibinfo{pages}{2531--2540}.
%Type = Inproceedings
\bibitem[{Garc\'ia Seco~de Herrera et~al.(2020)Garc\'ia Seco~de Herrera,
  Savran~Kiziltepe, Chamberlain, Constantin, Demarty, Doctor, Ionescu, and
  Smeaton}]{GSC2020}
\bibinfo{author}{A.~Garc\'ia Seco~de Herrera},
  \bibinfo{author}{R.~Savran~Kiziltepe}, \bibinfo{author}{J.~Chamberlain},
  \bibinfo{author}{M.~G. Constantin}, \bibinfo{author}{C.-H. Demarty},
  \bibinfo{author}{F.~Doctor}, \bibinfo{author}{B.~Ionescu},
  \bibinfo{author}{A.~F. Smeaton},
\newblock \bibinfo{title}{Overview of {MediaEval} 2020 predicting media
  memorability task: What makes a video memorable?},
\newblock in: \bibinfo{booktitle}{{MediaEval} 2020}, volume
  \bibinfo{volume}{2882}, \bibinfo{publisher}{CEUR-WS.org},
  \bibinfo{year}{2020}.
%Type = Inproceedings
\bibitem[{Krizhevsky et~al.(2012)Krizhevsky, Sutskever, and
  Hinton}]{krizhevsky2012imagenet}
\bibinfo{author}{A.~Krizhevsky}, \bibinfo{author}{I.~Sutskever},
  \bibinfo{author}{G.~E. Hinton},
\newblock \bibinfo{title}{Imagenet classification with deep convolutional
  neural networks},
\newblock in: \bibinfo{booktitle}{Advances in Neural Information Processing
  Systems}, \bibinfo{year}{2012}, pp. \bibinfo{pages}{1097--1105}.
%Type = Inproceedings
\bibitem[{Dalal and Triggs(2005)}]{dalal2005histograms}
\bibinfo{author}{N.~Dalal}, \bibinfo{author}{B.~Triggs},
\newblock \bibinfo{title}{Histograms of oriented gradients for human
  detection},
\newblock in: \bibinfo{booktitle}{2005 {IEEE} Computer Society Conference on
  Computer Vision and Pattern Recognition ({CVPR'05})},
  volume~\bibinfo{volume}{1}, \bibinfo{organization}{IEEE},
  \bibinfo{year}{2005}, pp. \bibinfo{pages}{886--893}.
%Type = Article
\bibitem[{Ojala et~al.(2002)Ojala, Pietikainen, and
  Maenpaa}]{ojala2002multiresolution}
\bibinfo{author}{T.~Ojala}, \bibinfo{author}{M.~Pietikainen},
  \bibinfo{author}{T.~Maenpaa},
\newblock \bibinfo{title}{Multiresolution gray-scale and rotation invariant
  texture classification with local binary patterns},
\newblock \bibinfo{journal}{IEEE Transactions on pattern analysis and machine
  intelligence} \bibinfo{volume}{24} (\bibinfo{year}{2002})
  \bibinfo{pages}{971--987}.
%Type = Inproceedings
\bibitem[{Simonyan and Zisserman(2015)}]{simonyan2014very}
\bibinfo{author}{K.~Simonyan}, \bibinfo{author}{A.~Zisserman},
\newblock \bibinfo{title}{Very deep convolutional networks for large-scale
  image recognition},
\newblock in: \bibinfo{booktitle}{International Conference on Learning
  Representations}, \bibinfo{year}{2015}.
%Type = Inproceedings
\bibitem[{Tran et~al.(2015)Tran, Bourdev, Fergus, Torresani, and
  Paluri}]{tran2015learning}
\bibinfo{author}{D.~Tran}, \bibinfo{author}{L.~Bourdev},
  \bibinfo{author}{R.~Fergus}, \bibinfo{author}{L.~Torresani},
  \bibinfo{author}{M.~Paluri},
\newblock \bibinfo{title}{Learning spatiotemporal features with 3d
  convolutional networks},
\newblock in: \bibinfo{booktitle}{Proceedings of the {IEEE} International
  Conference on Computer Vision}, \bibinfo{year}{2015}, pp.
  \bibinfo{pages}{4489--4497}.
%Type = Article
\bibitem[{De~Leeuw(2015)}]{deLeeuw_2015_jspsych}
\bibinfo{author}{J.~R. De~Leeuw},
\newblock \bibinfo{title}{jspsych: A javascript library for creating behavioral
  experiments in a web browser},
\newblock \bibinfo{journal}{Behavior research methods} \bibinfo{volume}{47}
  (\bibinfo{year}{2015}) \bibinfo{pages}{1--12}.

\end{thebibliography}
\end{document}